\pdfoutput=1
\documentclass[11pt]{article}
\usepackage{tikz}
\usetikzlibrary{calc}

\usepackage[preprint]{acl}

\usepackage{times}
\usepackage{latexsym}

\usepackage[T1]{fontenc}
\usepackage[utf8]{inputenc}
\usepackage{microtype}
\usepackage{inconsolata}
\usepackage{pifont}
\usepackage[table]{xcolor}
\usepackage{amsmath}
\newcommand{\E}{\mathbb{E}}
\newcommand{\cmark}{\ding{51}}
\newcommand{\xmark}{\ding{55}}
\usetikzlibrary{positioning,arrows.meta,shapes.geometric}

\usepackage{graphicx}
\usepackage{caption}
\usepackage{subcaption}
\usepackage{hyperref}
\usepackage{soul}
\usepackage{booktabs} 
\usepackage{amsmath}
\usepackage{pifont}
\usepackage{amssymb}
\usepackage{mathtools}
\usepackage{amsthm}
\usepackage{multirow}
\usepackage[capitalize,noabbrev]{cleveref}
\usepackage[textsize=tiny]{todonotes}
\usepackage{float}
\usepackage{enumitem}
\newtheorem{definition}{Definition}[section]

\setul{0.5ex}{0.3ex}
\setulcolor{blue}

\title{Detecting LLM Hallucination Through Layer-wise Information Deficiency: Analysis of Ambiguous Prompts and Unanswerable Questions}

\author{Hazel Kim$^\dagger$, Tom A. Lamb, Adel Bibi, Philip Torr$^\ddagger$, Yarin Gal$^\ddagger$ \\
        University of Oxford \\ \texttt{\{hazel.kim, yarin.gal\}@cs.ox.ac.uk}, \\
        \texttt{\{thomas.lamb, adel.bibi, philip.torr\}@eng.ox.ac.uk }}

\begin{document}
\maketitle
\def\thefootnote{$\dagger$}\footnotetext{Correspondence to Hazel Kim <hazel.kim@cs.ox.ac.uk>.}\def\thefootnote{\arabic{footnote}}
\def\thefootnote{$\ddagger$}\footnotetext{Equal advising.}\def\thefootnote{\arabic{footnote}}
\begin{abstract}
Large language models (LLMs) frequently generate confident yet inaccurate responses, introducing significant risks for deployment in safety-critical domains.
We present a novel, test-time approach to detecting model hallucination through systematic analysis of information flow across model layers. We target cases when LLMs process inputs with ambiguous or insufficient context. Our investigation reveals that hallucination manifests as usable information deficiencies in inter-layer transmissions. 
While existing approaches primarily focus on final-layer output analysis, we demonstrate that tracking cross-layer information dynamics ($\mathcal{L}$I) provides robust indicators of model reliability, accounting for both information gain and loss during computation. 
$\mathcal{L}$I integrates easily with pretrained LLMs without requiring additional training or architectural modifications.
\end{abstract}

\section{Introduction}

\begin{figure}[t]
    \centering
    \begin{subfigure}[b]{0.4\textwidth}
    \includegraphics[width=0.9\textwidth]{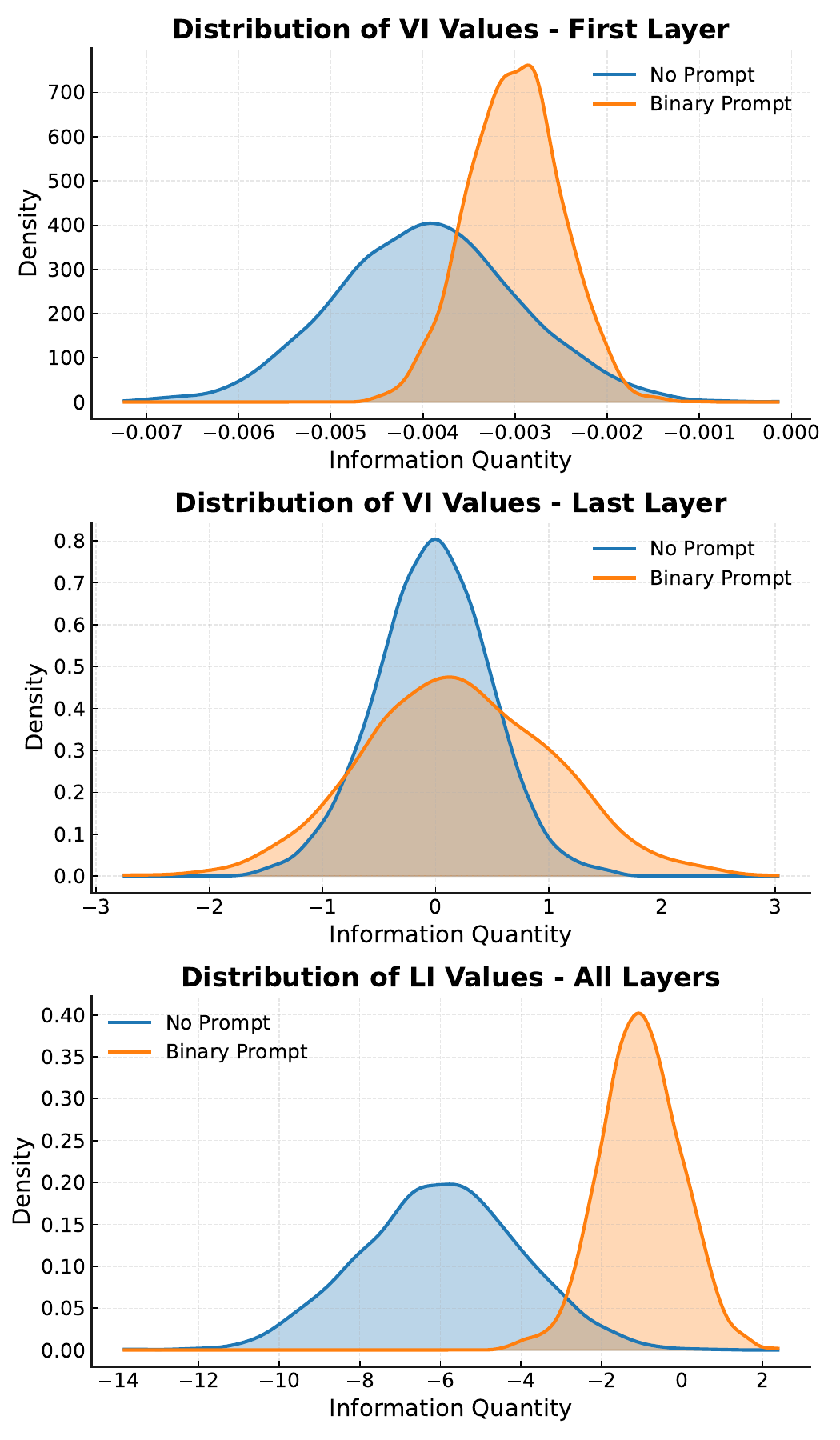}
    \end{subfigure}
    \vspace{-0.5em}
    \caption{Distribution of $\mathcal{V}$-information ($\mathcal{V}$I) values in first and last layers, and $\mathcal{L}$-information ($\mathcal{L}$I) values (summation of $\mathcal{V}$I scores across all layers), as a function of prompt ambiguity. Results compare two prompt categories: (1) no instruction prompts and (2) binary instruction prompts ('Is this answerable?').}
    \label{fig:VI_LI_comparison}
    \vspace{-2em}
\end{figure}

Large language models (LLMs) have achieved unprecedented success across diverse natural language tasks, particularly in complex reasoning ranging from commonsense to arithmetic knowledge~\citep{achiam2023gpt, touvron2023llama, abdin2024phi}. 
However, these models face a critical challenge known as \textit{hallucination}, a phenomenon where responses appear convincingly authoritative despite being inaccurate~\citep{JiLFYSXIBMF23, abs-2401-11817, abs-2308-05374}. 
While numerous empirical studies have investigated potential sources of hallucination, recent theoretical work by \citet{xu2024hallucination} demonstrates the fundamental impossibility of eliminating this issue through any computable function.

Following \citet{xu2024hallucination}, hallucination can be formally defined as the failure of LLMs to accurately reproduce the desired output of a computable function. This theoretical framework establishes that hallucination is an inherent characteristic of LLMs, persisting regardless of architectural choices, learning algorithms, prompting strategies, or training data composition. Building on this theoretical foundation, we hypothesize that hallucination emerges when LLMs lack sufficient information necessary for their computational functions to transmit messages across their internal processing systems effectively.

\begin{figure*}[t!]
    \centering
    \vspace{-0.5em}
    \includegraphics[width=0.97\linewidth]{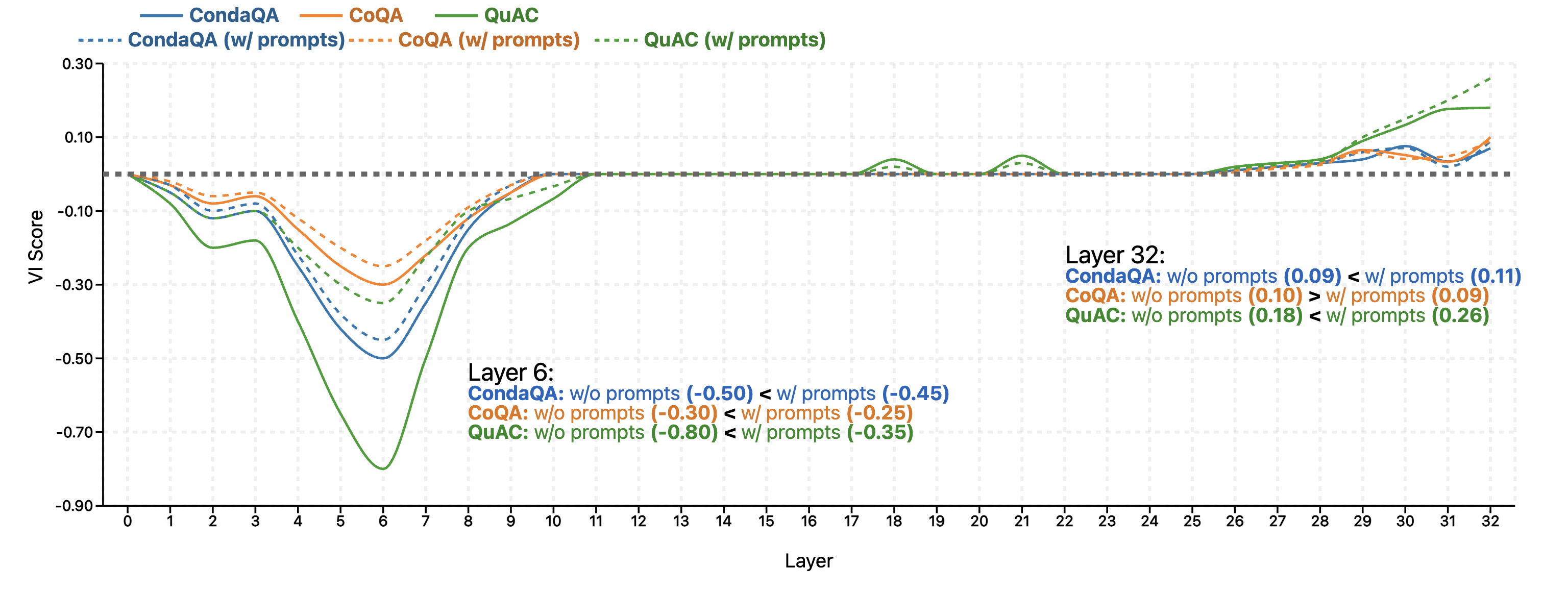}
    \vspace{-1.1em}
    \caption{$\mathcal{V}$-usable information ($\mathcal{V}$I; \citet{XuZSSE20}) across layers for CondaQA, CoQA, and QuAC. 
    Solid lines denote models without instruction prompts and dashed lines denote models with binary prompts ("Is this answerable?"). 
    The quantity of $\mathcal{V}$I is not monotonically increased or decreased across depth.
    By the final layer (32), the relative effect of prompts becomes inconsistent across datasets, highlighting the value of analyzing all layers.}
    \label{fig:VI_info_layers}
    \vspace{-1em}
\end{figure*}

Prior research has focused primarily on analyzing final outputs to assess LLM confidence~\citep{osband2023epistemic, AhdritzQ0BE24, LinT024} or identifying inherent data ambiguities that lead to predictive uncertainty~\citep{ColeZGEDE23, KuhnGF23}. 
However, the fundamental internal mechanisms of LLMs remain under-explored. 
Our analysis reveals that uncertainty estimation based solely on output layers or final computations overlooks critical insights into model \textit{self-confidence}, thereby limiting our ability to detect hallucinatory behaviors.

To investigate LLM internal mechanisms, we build upon recent information theory frameworks proposed by~\citet{XuZSSE20} and~\citet{EthayarajhCS22}. Their work introduces the concept of $\mathcal{V}$-usable information---the quantity of information a model family $\mathcal{V}$ can utilize to predict $Y$ given $X$. This metric indicates prediction difficulty: lower $\mathcal{V}$-usable information corresponds to more challenging predictions for $\mathcal{V}$.

While these findings are significant, we empirically demonstrate that $\mathcal{V}$-usable information provides sub-optimal insight into model self-confidence because it only applies to the final layer. 
We propose layer-wise usable information ($\mathcal{L}$I), which quantifies the information changes within certain layers and aggregates those information dynamics across all model layers. As shown in Fig.~\ref{fig:VI_LI_comparison}, $\mathcal{L}$I provides more reliable indicators of LLM performance than final layer $\mathcal{V}$-usable information ($\mathcal{V}$I), particularly in detecting subtle variations in instruction prompt effectiveness.

Prior work on $\mathcal{V}$-usable information established that computation can \textit{create} usable information during feature extraction, constituting a violation of the data processing inequality (DPI) in information theory~\citep{lecun2015deep, XuZSSE20}. Our research extends this understanding by demonstrating that LLMs both \textit{create} and \textit{lose} usable information during layer-wise updates. 
As illustrated in Fig.\ref{fig:VI_info_layers}, information flow is non-monotonic across layers, highlighting the limitation of analyzing only the final layer's computation.

To evaluate our framework, we focus on scenarios where LLMs must respond to queries with instruction prompts of different ambiguities (i.e., \textit{ambiguous prompts}) and with constrained contextual information (i.e., \textit{unanswerable} questions), settings particularly prone to hallucination. Our case study in Table~\ref{table:case_study} demonstrates that $\mathcal{L}$I strongly correlates with the difficulty of the question, influenced by the answerability and the clarity of the prompt, unlike $\mathcal{V}$I that shows no significant correlation with the model prediction.

\definecolor{lightblue}{rgb}{0.85,0.93,1}
\definecolor{lightgray}{rgb}{0.9,0.9,0.9}

\renewcommand{\arraystretch}{1.2}
\begin{table*}[t!]
    \centering
    \scriptsize
    \arrayrulecolor{gray}
    \setlength{\arrayrulewidth}{0.5mm}
    \setlength{\tabcolsep}{6pt}
    \begin{tabular}{p{15cm}}
    \toprule
    \rowcolor{lightgray}
    \textbf{Context} \\
    \midrule
     \textcolor{blue}{\textbf{Once there was a beautiful fish named Asta.}} Asta lived in the ocean. There were lots of other fish in the ocean where Asta lived. They played all day long. One day, a bottle floated by over the heads of Asta and his friends. They looked up and saw the bottle. ... They took the note to Asta's papa. "What does it say?" they asked. Asta's papa read the note. He told Asta and Sharkie, "This note is from a little girl. She wants to be your friend. If you want to be her friend, we can write a note to her. \textcolor{blue}{\textbf{But you have to find another bottle so we can send it to her." And that is what they did.}} \\ 
    \bottomrule
    \end{tabular}
    \begin{tabular}{p{5cm} p{3.8cm} c c c c c}
        \rowcolor{lightgray}
        \toprule
        \textbf{Instruction Prompt} & \textbf{Question} & \textbf{Ground-Truth Label} & \textbf{Prediction} & \textbf{\textit{VI}} & \textbf{\textit{LI}} \\
        \midrule
        \textbf{Binary: ``Is this answerable?''} & What was the name of the fish? & Yes. & \checkmark & 0.3 & 0.2 \\
            & Were they excited? & No. & \checkmark & 0.3 & -0.4 \\ \midrule
        \textbf{Open-ended Prompt:  } & What was the name of the fish? & Asta. & \cmark & 0.3 & -0.7 \\
        \textbf{``Answer the question or say don\'t know''} & Were they excited? & Don’t know. & \xmark & -0.3 & -1.4 \\ \midrule
        \textbf{No prompts} & What was the name of the fish? & Asta. & \checkmark & 1.2 & -6.8 \\
        & Were they excited? & Unknown. & \xmark & 0.2 & -7.7 \\
        \bottomrule
    \end{tabular}
    \caption{The \textit{$\mathcal{L}$I} scores provide a more comprehensive overview of prompt ambiguity compared to \textit{$\mathcal{V}$I}. \textbf{\textit{Prediction}:} prediction generated by a language model --- \checkmark: correct, \xmark: incorrect. \textit{$\mathcal{V}$I}: $\mathcal{V}$-usable information only applied to the final layer~\citep{XuZSSE20}. \textit{$\mathcal{L}$I}: Layer-wise usable information accumulated across layers (our proposed method).}
    \vspace{-1.5em}
    \label{table:case_study}
\end{table*}

\paragraph{Contributions} Our primary contributions are:

\begin{itemize}[itemsep=0.5ex]
    \item We propose $\mathcal{L}$I as a superior detector of unanswerable questions compared to existing baselines (Section~\ref{section:unanswerability}), without requiring architectural modifications or additional training.
    \item We demonstrate that $\mathcal{L}$I effectively captures model confidence across varying levels of task difficulty induced by different instruction prompts (Section~\ref{section:baselines}).
    \item We interpret that comprehensive layer tracking provides better insights into model internal confidence than single-layer analysis, using either initial or final layer (Sections~\ref{section:3} and~\ref{section:all_layers}). 
\end{itemize}

\section{Related Work}

\paragraph{Contextual vs Factual Hallucinations}
Large language models (LLMs) often generate inaccurate outputs despite having access to correct information in their input context. This phenomenon is known as contextual hallucination ~\citep{chuang-etal-2024-lookback}. 
This issue is particularly concerning in high-stakes domains such as medicine and law, where acknowledging information gaps is preferable to making unfounded assumptions.

Most prior studies, however, focus on fact-based hallucination arising from parametric knowledge without input context. 
These hallucinations may result from inherent learning limitations or training data deficiencies, making their root causes difficult to isolate. 
Existing approaches have detected and mitigated such errors by using substantial annotated data to analyze various model components, including hidden states~\citep{burns2022discovering, azaria2023internal}, MLP and attention block outputs~\citep{zhang2024truthx, simhi2024constructing}, and attention head outputs~\citep{li2023inference, simhi2024constructing, chen2024truth}. 

In contrast, contextual hallucination remains comparatively understudied. Existing work in this area has so far relied on annotated datapoints~\citep{chuang-etal-2024-lookback} without probing the model internal mechanisms. Our research addresses this gap by examining contextual hallucination as an ideal setting to explore how LLMs behave when faced with insufficient information. We deliberately place models in situations where they must respond despite clearly inadequate input information, enabling us to study the fundamental nature of LLM hallucination behavior.

\paragraph{Unanswerable Questions}

Existing work has analyzed the model’s capability to detect unanswerable questions from three main perspectives. One is self-evaluation which allows language models to generate probabilistic scores of how much the models believe their answers are trustworthy~\citep{Kadavath22, YinSGWQH23} as we get access to advanced, well-calibrated models that can generate reliable results. The second perspective involves identifying the subspace of the model that is specifically responsible for answerability~\citep{SlobodkinGCDR23}. The third approach uses label information to train LLMs on whether questions are answerable, employing methods such as instruction-tuning or calibration~\citep{JiangADN21, kapoorGRPDGW2024}. While these studies suggest that LLMs can learn to express their confidence in responses when provided with additional information, they rely heavily on external calibration or fine-tuning. The outcome depends on the quality of the additional information or that of the annotation work. Unlike prior work, our investigation does not require label annotations to fine-tune classifiers or calibration tools to detect model confidence in their generated answers or ambiguous, unanswerable questions. We aim to obtain a computationally feasible method that applies to universal large language models.

\label{section:2}
\paragraph{Model Usable Information}

Our analysis builds upon the information-theoretic framework introduced by \citet{XuZSSE20} and expanded by \citet{EthayarajhCS22}, focusing on quantifying the "usable information" accessible to models. Given a model family $\mathcal{V}$ that maps inputs $X$ to outputs $Y$, the concept of $\mathcal{V}$-usable information measures how effectively $\mathcal{V}$ can leverage input data to predict outputs. Lower usable information correlates with increased prediction difficulty. For instance, encrypted or linguistically complex inputs reduce $\mathcal{V}$-usable information, increasing predictive challenges within the same model family.

This approach extends traditional information theory, particularly Shannon's mutual information~\citep{shannon} and the data processing inequality (DPI;~\citealp{pippenger1988reliable}). 
Mutual information quantifies the theoretical information shared between inputs and outputs, while DPI states that this quantity cannot increase as data passes through further transformations. 
Although these measures describe theoretical information flow, they do not capture how much of that information is practically usable by a given model family. 
In practice, usable information can diverge from the classical measures for two reasons:
(1) computational constraints limit the extent to which models can realize the ideal mapping from $X$ to $Y$~\citep{XuZSSE20}; and 
(2) deep representation learning not only restructures inputs across layers to extract features from incomplete or noisy data, but also leverages prior knowledge stored in pretrained weights~\citep{lecun2015deep, goldfeld2021sliced}. 
These considerations motivate the notion of \(\mathcal{V}\)-usable information, which explicitly quantifies the information that is practically available to a given model family.
The two frameworks leveraging $\mathcal{V}$-usable information are:

\textbf{Predictive $\mathcal{V}$-information} quantifies aggregate informativeness or dataset difficulty given model family constraints, expressed as $I_{\mathcal{V}}(X \rightarrow Y)$~\citep{XuZSSE20}.

\textbf{Pointwise $\mathcal{V}$-information} evaluates the information usability of individual instances relative to a specific dataset distribution, denoted as $\text{PVI}(x \rightarrow y)$~\citep{EthayarajhCS22}.

Formally, predictive $\mathcal{V}$-information is defined as:
\begin{definition}[Predictive $\mathcal{V}$-information, \citealt{XuZSSE20}]
Given predictive conditional entropy $H_{\mathcal{V}}(Y|X)$:
\begin{equation}
I_{\mathcal{V}}(X\rightarrow Y) = H_{\mathcal{V}}(Y|\varnothing) - H_{\mathcal{V}}(Y|X).
\end{equation}
\end{definition}

Traditionally, the model family $\mathcal{V}$ has been instantiated as supervised models such as BERT~\citep{devlin2019bert}, trained to minimize expected log-loss risk on labeled datasets $(x,y) \sim p_{\mathcal{D}}$. 
This yields the following definitions of conditional predictive entropy:
\begin{align*}
H_{\mathcal{V}}(Y|X) &= \mathbb{E}_{(x,y)\sim p_{\mathcal{D}}}\left[-\log_2 p(y|x)\right],\\
H_{\mathcal{V}}(Y|\varnothing) &= \mathbb{E}_{(x,y)\sim p_{\mathcal{D}}}\left[-\log_2 p(y|\varnothing)\right].
\end{align*}

In contrast to these supervised approaches, our proposed $\mathcal{L}I$ does not require training. 
In the next section, we will explain our proposed methodology. Detailed background on $\mathcal{V}$-usable information is provided in Appendix~\ref{appendix_sec:usable_info}.

\section{Layer-Wise Usable Information}
\label{section:3}

\setcounter{algorithm}{0}
\begin{algorithm}[t]
\footnotesize
\caption{Computing layer-wise usable information ($\mathcal{L}I$) without fine-tuning}
\textbf{Input:} dataset $\mathcal{D} = \{(c_i, q_i)\}_{i=1}^m$, pretrained model with layers $\mathcal{L}$ \\
\textbf{Output:} $\mathcal{L}I(C \to Q)$
\begin{algorithmic}[1]

\STATE $\varnothing \gets$ empty context (null string)

\FOR{each example $(c_i, q_i) \in \mathcal{D}$}
  \FOR{each layer $\ell \in \mathcal{L}$}
    \STATE Compute token log-probs $p_\ell(q_t \mid q_{<t}, \varnothing)$
    \STATE Compute token log-probs $p_\ell(q_t \mid q_{<t}, c_i)$
    \STATE $H_{\ell}^{(i)}(Q \mid \varnothing) \gets \frac{1}{T_i}\sum_{t=1}^{T_i} -\log_2 p_\ell(q_t \mid q_{<t}, \varnothing)$
    \STATE $H_{\ell}^{(i)}(Q \mid C) \gets \frac{1}{T_i}\sum_{t=1}^{T_i} -\log_2 p_\ell(q_t \mid q_{<t}, c_i)$
    \STATE $I_{\ell}^{(i)} \gets H_{\ell}^{(i)}(Q \mid \varnothing) - H_{\ell}^{(i)}(Q \mid C)$
  \ENDFOR
\ENDFOR

\STATE $\mathcal{L}I(C \to Q) \gets \tfrac{1}{m} \sum_{i=1}^m \sum_{\ell \in \mathcal{L}} I_{\ell}^{(i)}$

\end{algorithmic}
\label{algo:li}
\end{algorithm}

\begin{figure}
\centering
\begin{tikzpicture}[
  box/.style={
    draw,
    fill=blue!5,
    thick,
    rounded corners,
    text width=2cm,
    align=center,
    minimum height=0.6cm,
    inner sep=2pt,
    font=\scriptsize
  },
  arrow/.style={
    -{Stealth[length=2mm,width=1mm]},
    ultra thin
  }
]

  \node[box] (start) {For each\\$(c_i,q_i)$};

  \node[box, below left=0.6cm and 0.4cm of start]  (A)
    {Path A:\\No Context\\$p(q\mid\varnothing)$};
  \node[box, below right=0.6cm and 0.4cm of start] (B)
    {Path B:\\With Context\\$p(q\mid c_i)$};

  \node[box, below=0.8cm of A] (H0)
    {$H_\ell(Q\mid\varnothing)$\\
     $=-\E[\log $ \\
     $p(q_t\mid q_{<t},\varnothing)]$};
  \node[box, below=0.8cm of B] (HC)
    {$H_\ell(Q\mid C)$\\
     $=-\E[\log$ \\
     $p(q_t\mid q_{<t},c_i)]$};

  \node[box, below=1cm of $(H0)!0.5!(HC)$] (D)
    {$I_\ell(c_i\!\to\!q_i)$\\
     $=H_\ell(Q\mid\varnothing)$\\
     $\;-\;H_\ell(Q\mid C)$};
     
  \draw[arrow] (start) -- (A);
  \draw[arrow] (start) -- (B);
  \draw[arrow] (A) -- (H0);
  \draw[arrow] (B) -- (HC);
  \draw[arrow] (H0.south) |- (D.west);
  \draw[arrow] (HC.south) |- (D.east);

\end{tikzpicture}
\caption{Illustration of computing layer-wise usable information for an example $(c_i,q_i)$ at a \emph{single layer} $\ell$.}
\label{fig:illustration}
\vspace{-1em}
\end{figure}

We extend the $\mathcal{V}$-usable information framework to quantify information at the layer level in generative language models. 
\emph{Layer-wise usable information} ($\mathcal{L}$I) measures how much a context $C$ changes the predictive entropy of a question $Q$ at each layer $\ell$. 
The per-layer contribution is $I_\ell$, and the total $\mathcal{L}I = \sum_{\ell \in \mathcal{L}} I_\ell$ aggregates these differences across all layers.

Concretely, given a context $C$, the model generates a free-form answer to a question $Q$. 
Let $\mathcal{L}$ denote the set of layers in a pre-trained language model. 
Each layer $\ell \in \mathcal{L}$ produces hidden representations. 
Projected through the pretrained language model head, the hidden states at layer $\ell$ induce a conditional distribution
\[
f^{(\ell)}:\ \mathcal{C}\cup\{\varnothing\} \;\to\; P(\mathcal{Q}),
\]
where $\mathcal{Q}$ is the token vocabulary. 
For a given question prefix $q_{<t}$ and context $c\in\mathcal{C}$ (or $\varnothing$), this distribution specifies probabilities $p_\ell(q_t \mid q_{<t}, c)$ over the next token $q_t \in \mathcal{Q}$.

This formulation allows us to measure usable information both at the level of individual layers, $I_\ell$, and in aggregate across the model, $\mathcal{L}I$. Unlike prior work, we do not fine-tune $f^{(\ell)}$ on labeled data, but instead directly use the pretrained model outputs.

\paragraph{Definition 3.1 (Predictive conditional $\ell$-entropy).} 
Let $q_t$ denote the $t$-th token in a question sequence $q \in Q$. 
The predictive conditional entropy at layer $\ell$ is
\begin{equation}
H_{\ell}(Q|C) = \mathbb{E}_{q \sim Q}\Big[-\log_2 p_\ell(q_t \mid q_{<t}, C)\Big],
\end{equation}
and, similarly for the null context,
\begin{equation*}
H_{\ell}(Q|\varnothing) = \mathbb{E}_{q \sim Q}\Big[-\log_2 p_\ell(q_t \mid q_{<t}, \varnothing)\Big].
\end{equation*}
These quantities represent the predictive uncertainty of the distributions derived from layer $\ell$, either conditioned on the context $C$ or without it. 
In practice, we report per-token entropies by averaging these values across all positions $t$ in the question sequence.

\paragraph{Definition 3.2 (Predictive $\mathcal{L}$-information).} 
The layer-wise usable information from $C$ to $Q$ is defined as
\begin{equation}
\begin{aligned}
\mathcal{L}I(c \to q) &= \sum_{\ell \in \mathcal{L}} I_\ell(c \to q), \\
I_\ell(c \to q) &= H_\ell(Q|\varnothing) - H_\ell(Q|C).
\end{aligned}
\end{equation}
Here, $I_\ell(c \to q)$ measures the change in entropy due to the presence of context at layer $\ell$, and $\mathcal{L}I(c \to q)$ aggregates these contributions across all layers. 
Algorithm~\ref{algo:li} and Figure~\ref{fig:illustration} illustrate this computation.

\subsection{Implications}
\begin{figure*}[t]
    \centering
    \includegraphics[width=0.95\linewidth]{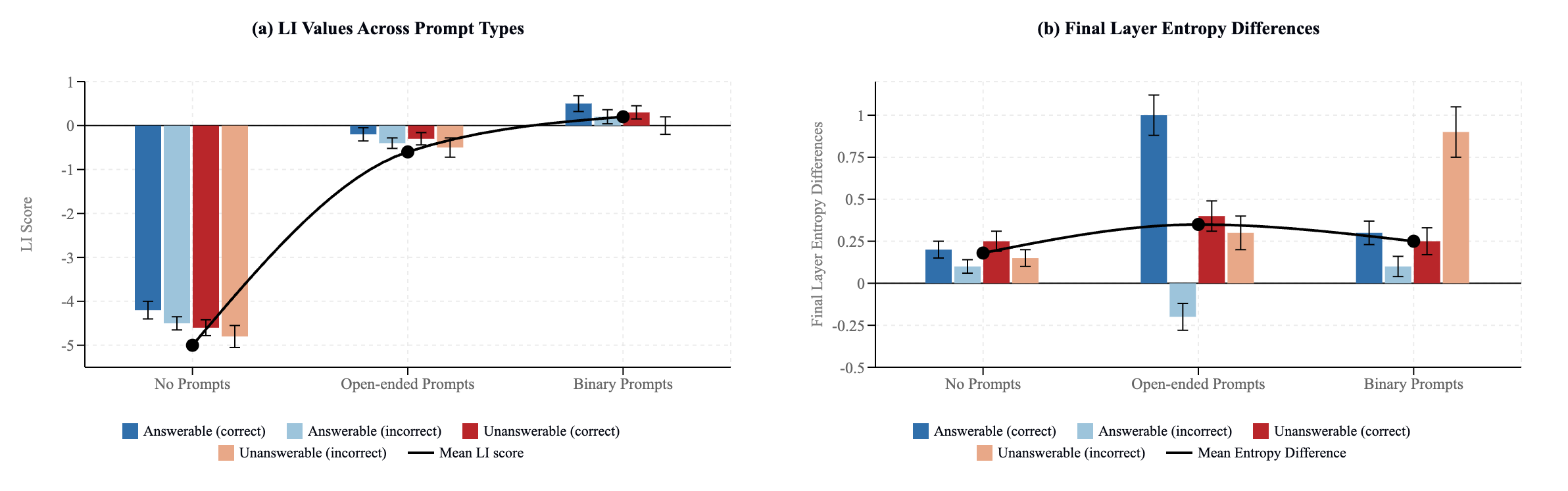}
    \caption{Impact of instruction prompts on layer information in QuAC. 
    (a) $\mathcal{L}$I: scores increase systematically as prompts become more explicit (no prompt $\rightarrow$ open-ended $\rightarrow$ binary). Within each prompt type, correct answers have higher scores than incorrect ones (answerable-correct $>$ answerable-incorrect; unanswerable-correct $>$ unanswerable-incorrect). (b) Final-layer $\mathcal{V}$I: scores show no consistent progression and correct–incorrect separation.}
    \label{fig:error_different_prompts}
    \vspace{-1em}
\end{figure*}

Using layer-wise usable information ($\mathcal{L}$I), we contribute to the following accomplishments:

\begin{itemize}
    \item Detection for unanswerable questions by computing $\mathcal{L}I(c \rightarrow q)$ in datasets \{$c\in C, q \in Q$\} for the same $\mathcal{L}$: we classify questions that lack sufficient usable information as unanswerable, likely to be inaccurate responses (Fig.~\ref{fig:performance1} and~\ref{fig:performance2}). 
    \item  Evaluation on different prompts with $Q$ for $\mathcal{L}$ by estimating $\mathcal{L}I(C \rightarrow Q')$.  We quantify how different instruction prompt $Q'$ influences usable information (Figs.~\ref{fig:error_different_prompts} and Table~\ref{tab:li_scores_random_prompts}).
    \item Analysis of importance of all layer information estimating $\mathcal{L}I(C \rightarrow Q)$. Aggregating across layers shows that full $\mathcal{L}I$ provides stronger separation between answerable and unanswerable questions than any single layer alone (Figs.~\ref{fig:error_different_prompts} and~\ref{fig:cumulative_vi} and Table~\ref{tab:fact_hallucination}).
\end{itemize}

\section{Experiments}

We demonstrate that layer-wise usable information ($\mathcal{L}$I) is an effective way to capture the ambiguity of prompts and detect unanswerable questions for large language models. 

\subsection{Experimental Setup}

\textbf{Evaluation Metric.}
We classify unanswerable questions based on uncertainty scores.
The ground-truth labels of the unanswerability based on the contextual information are provided by the original benchmark datasets~\citep{ReddyCM19, ChoiHIYYCLZ18, Ravichander0M22}.
We evaluate uncertainty under the assumption that we assess whether to trust a model's generated response in a given context, i.e., deciding whether to accept an answer to a question. 
Our primary metric for this assessment is the area under the receiver operating characteristic curve (AUROC).
The AUROC measures the \emph{discrimination} ability of a scoring function---how well it separates correct from incorrect predictions. 
In our setting, this corresponds to distinguishing answerable from unanswerable questions using the uncertainty score provided by $\mathcal{L}I$.
Higher AUROC scores indicate better performance, with a perfect score of 1 representing optimal uncertainty estimation, while a score of 0.5 represents random uncertainty.

We choose to use the AUROC as it suits well for evaluating uncertainty in free-form text responses, as opposed to calibration measures like the Brier score, frequently used in classification tasks or multiple-choice question answering. The Brier score requires calculating the total probability mass assigned to all possible tokens of a correct answer sequences. This causes the task to be intractable in free-form text settings where probabilities with respect to meanings are unavailable. 
Therefore we use the AUROC to capture the uncertainty associated with the model's outputs more accurately, and classify the unanswerable questions. One exception is model answers that we simply match the lexical words by instructing models to generate "Yes" or "No" to additional question prompts asking if they can answer the question based on the context. 

\textbf{Baselines.} We evaluate our method against several benchmarks, including model-generated answers, \textsc{P(True)}~\citep{kadavath2022language}, predictive token entropy and normalized entropy~\citep{malinin2020uncertainty}, semantic entropy~\citep{farquhar2024detecting}, and pointwise $\mathcal{V}$-information (PVI)~\citep{EthayarajhCS22} on the first and the last layers respectively. 
Model-generated answers are raw responses by models. 
\textsc{P(True)} measures the probability that a LLMs predict the next token as ‘True’ when provided with few-shot prompts that compare a primary answer to various alternative answers.
Predictive entropy is calculated by conditional entropy over the output distribution. 
Predictive normalized entropy is obtained by dividing the total sequence-level entropy, computed as the negative log-likelihood, by the sequence length.
We use a single model to meausre the normalized entropy, following the setups by~\citep{kuhn2023semantic}. Semantic entropy follows the confabulation mechanism to classify unanswerable questions. PVI is to measure difficult datapoints. We assume that difficult instances for language models are likely to be unanswerable questions.

\textbf{Models.} We use Llama3~\citep{dubey2024llama} and Phi3 models~\citep{abdin2024phi}. We vary the size of the models between 3.8B, 8B, and 14B parameters. We report our headline results using the most computationally efficient model, with 3.8B parameters unless we notify otherwise. In all cases we use only a single unmodified model since recent foundation models are not practical to modify the architectures and are often too costly to fine-tune them on datasets. Above all, we are interested in investigating internal language model behaviors than simply achieving optimal performance results. Hence we use them in their pre-trained form.

\textbf{Datasets.}
We use Conversational Question Answering Challenge dataset (CoQA)~\citep{ReddyCM19}  and Question Answering In Context (QuAC)~\citep{ChoiHIYYCLZ18} as question-answering tasks, where the model responds to questions using information from a supporting context paragraph. Our experiments are conducted on the development set, which contains approximately 8,000 questions.
We also use CondaQA~\citep{Ravichander0M22} which features 14,182 question-answer pairs with over 200 unique negation cues in addition to CoQA and QuAC to evaluate how trustworthy the $\mathcal{L}$I is to detect unanswerable questions. Given that goal, we employ a 1-to-1 ratio of answerable to unanswerable questions for a clear performance evaluation.

\subsection{Do $\mathcal{L}$I scores indicate the ambiguity of prompts?}
\label{section:baselines}

\begin{table}[t!]
    \centering
    \scriptsize
    \begin{tabular}{lccc}
        \toprule
        \textbf{Prompt} & \textbf{Ans.} & \textbf{Unans.} & \textbf{$\Delta$ (Ans.–Unans.)} \\
        \midrule
        \tiny Binary (``Is this question answerable?'') & 0.322 & 0.321 & 0.001\\
        \tiny Always answer YES. & 0.329 & 0.295 & 0.033 \\
        \tiny Always answer NO. & 0.316 & 0.287 & 0.029\\
        \tiny Is this question interesting? & 0.031 & -0.008 & 0.039\\
        \tiny Did your family like cappuccino? & 0.180 & 0.155 & 0.025 \\
        \tiny Can you give the wrong answer? & 0.134 & 0.089 & 0.044 \\
        \tiny Can you give the correct answer? & 0.180 & 0.117 & 0.064 \\
        \tiny Do you like your answer? & 0.161 & -0.023 & 0.184\\
        \bottomrule
    \end{tabular}
    \caption{$\mathcal{L}$I scores on QuAC (100 examples averaged). 
    Task-relevant binary prompts yield the highest scores on question examples, while irrelevant prompts reduce them. 
    Larger deltas ($\Delta$) indicate stronger separation of (un)answerability, which remains detectable even under irrelevant prompts.}
    \vspace{-1.5em}
    \label{tab:li_scores_random_prompts}
\end{table}
 
Figure~\ref{fig:error_different_prompts} compares how instruction prompts influence $\mathcal{L}$I and final-layer $\mathcal{V}$I. 
Without prompts, $\mathcal{L}$I scores remain strongly negative ($-4$ to $-5$), indicating high uncertainty. 
As prompts become more explicit, scores increase systematically: open-ended prompts yield intermediate values ($-0.5$ to $0$), while binary prompts produce the highest (slightly positive) scores. 
This progression shows that $\mathcal{L}$I is sensitive to the specificity of instructions, reliably reflecting prompt ambiguity.  
In contrast, final-layer $\mathcal{V}$I (Figure~\ref{fig:error_different_prompts}b) shows no consistent progression across prompt types. 
Because $\mathcal{V}$I was originally defined only for the final layer, such analyses obscure the consistent benefits of explicit prompts that are visible in intermediate representations. 

Table~\ref{tab:li_scores_random_prompts} further illustrates how prompt relevance affects $\mathcal{L}$I. 
With binary prompts, scores are around 0.32, higher than most random prompts. 
Answer-forcing prompts such as \textit{“Always answer YES/NO”} also yield relatively high $\mathcal{L}$I values, since these responses are easy for the model to generate. 
However, their scores remain below those of task-aligned prompts, reflecting that outputs are produced mechanically rather than through correct reasoning. 
Irrelevant prompts (\textit{“Is this question interesting?”}, \textit{“Did your family like cappuccino?”}) push scores substantially lower, showing how off-task instructions increase ambiguity. 
Misdirecting prompts (\textit{“Can you give the wrong answer?”}) reduce scores even further, consistent with the uncertainty introduced by conflicting instructions. 
By contrast, task-aligned prompts (\textit{“Can you give the correct answer?”}) partially restore $\mathcal{L}$I, while meta-reflective prompts (\textit{“Do you like your answer?”}) cause the strongest shifts, showing that self-assessment language accentuates the effect of prompt relevance.  

Overall, these patterns demonstrate that $\mathcal{L}$I is sensitive not only to the presence of an instruction but also to its relevance and quality. 
Irrelevant or adversarial prompts depress $\mathcal{L}$I, while task-relevant or self-reflective prompts elevate it, confirming that $\mathcal{L}$I provides a robust signal of prompt ambiguity.

\subsection{Do $\mathcal{L}$I scores capture unanswerable questions?}
\label{section:unanswerability}

\begin{figure}[t]
\centering
\begin{minipage}{\linewidth}
    \centering
    \includegraphics[width=\linewidth]{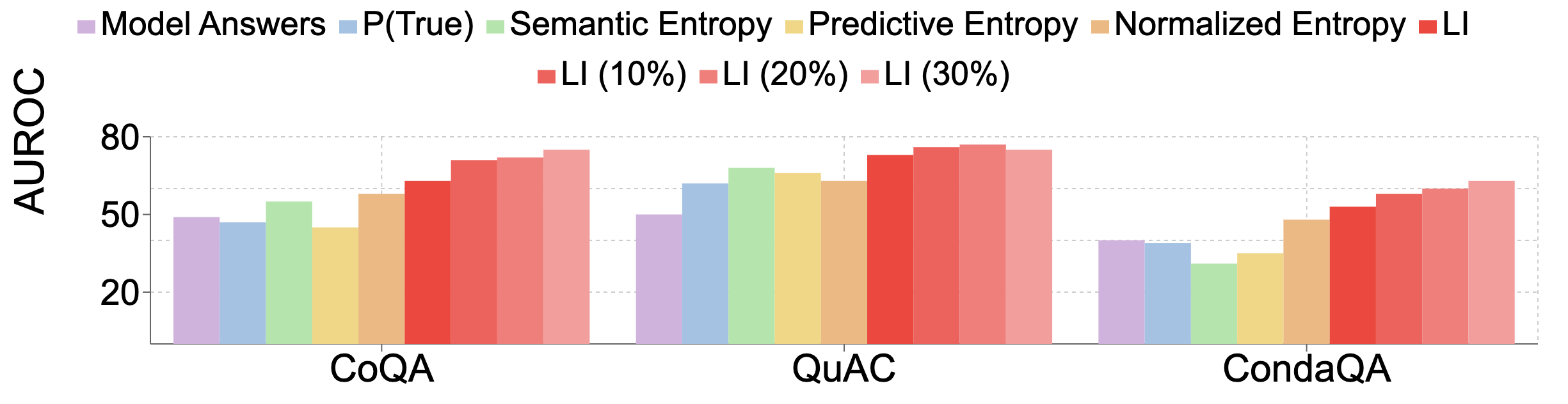}
    \caption{Performance of (un)answerability detection across datasets, comparing $\mathcal{L}$I with other baselines. $\mathcal{L}$I(10\%), $\mathcal{L}$I(20\%), and $\mathcal{L}$I(30\%) shows the rejection rate based on low scores.}
    \label{fig:performance1}
\end{minipage}

\vspace{1em}

\begin{minipage}{\linewidth}
    \centering
    \begin{subfigure}[b]{\linewidth}
        \includegraphics[width=\linewidth]{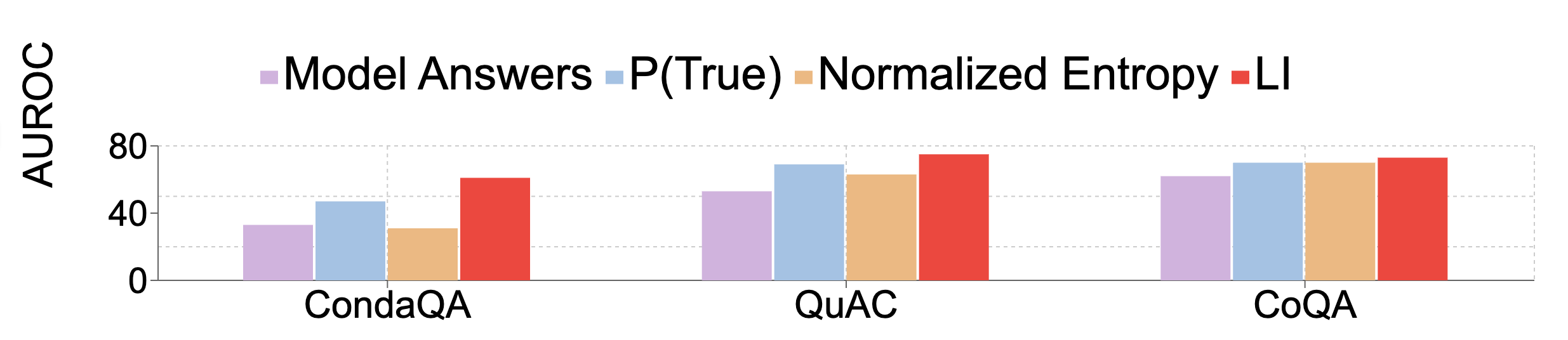}
        \caption{Phi3 medium (14B)}
    \end{subfigure}
    \begin{subfigure}[b]{\linewidth}
        \includegraphics[width=\linewidth]{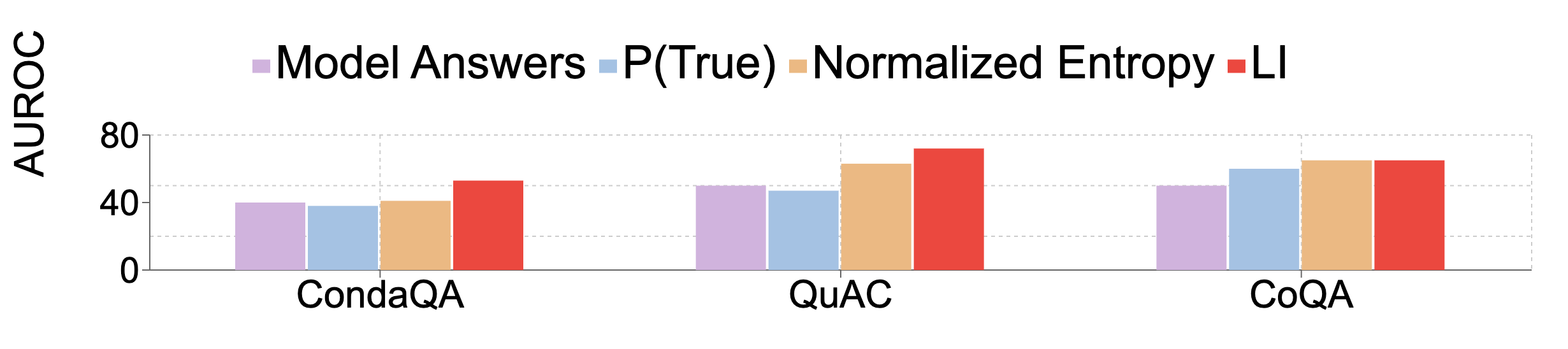}
        \caption{Llama3 (8B)}
    \end{subfigure}
    \vspace{-1.5em}
    \caption{Performance of (un)answerability detection, compared to selected competitive baselines.}
    \label{fig:performance2}
\end{minipage}
\end{figure}

\begin{table}[t]
    \centering
    \scriptsize
    \begin{tabular}{l|c|c}
        \toprule
         & \textbf{Answerable} & \textbf{Unanswerable} \\
        \midrule
        Correct & 0.4668  & 0.5088 \\
         & \tiny (e.g., Yes, there were clues...) & \tiny(e.g., I cannot provide ...) \\
         \midrule
        Incorrect & 0.0189  & -0.0578  \\
         &\tiny (e.g., In 1983,...) &  \tiny(e.g., Yes, she had...) \\
        \midrule
        Unsure & 0.0932  & -  \\
         & \tiny(e.g., I cannot provide...) &\tiny (LLMs are correct in this case.) \\
        \bottomrule
    \end{tabular}
    \caption{$\mathcal{L}$I scores for answerable and unanswerable questions with an \textit{instruction prompt: Are you certain about the answer?} on QuAC (100 examples averaged). Unsure indicates that models express the uncertainty.}
    \label{tab:certainty}
    \vspace{-2em}
\end{table}

Beyond prompt ambiguity, $\mathcal{L}$I also serves as a reliable signal for unanswerable questions. 
Across CoQA, QuAC, and CondaQA, $\mathcal{L}$I consistently outperforms baseline methods in distinguishing answerable from unanswerable questions (Figure~\ref{fig:performance1}). 
The advantage holds across different models and parameter sizes (Figure~\ref{fig:performance2}), whereas semantic entropy (SE) performs poorly on this task despite strong results elsewhere. 
This highlights that $\mathcal{L}$I is especially suited for (un)answerability detection, while other existing methods are not.  
Rejection analysis strengthens this conclusion. 
Filtering out predictions with the lowest $\mathcal{L}$I values steadily improves AUROC (Figure~\ref{fig:performance1}), confirming that low $\mathcal{L}$I reliably flags unanswerable cases. 

This trend extends more broadly across different prompt types in Table~\ref{tab:li_scores_random_prompts}. 
With task-relevant binary prompts, unanswerable questions tend to have slightly lower scores, indicating sensitivity to unanswerability. 
This pattern persists under random, irrelevant, or misleading prompts. Though the absolute values vary, unanswerable questions remain associated with relatively small $\mathcal{L}$I scores. 
This consistency demonstrates that $\mathcal{L}$I distinguishes answerable from unanswerable questions not only under optimal instructions, but also when the prompting conditions are weak or noisy.

Another noteworthy role of $\mathcal{L}$I appears in Table~\ref{tab:certainty}, where the explicit certainty prompt (\textit{“Are you certain about the answer?”}) elicits a different pattern from the binary prompt (\textit{“Is this question answerable?”}). 
Correct unanswerable responses achieve the highest $\mathcal{L}$I scores (0.509), even higher than correct answerable ones (0.467), reflecting that the model is appropriately certain about its own uncertainty. 
Incorrect answers, by contrast, yield the lowest values (0.019 for answerable, –0.058 for unanswerable), as expected from clear mismatches. 
Answerable questions where the model expressed uncertainty (“unsure”) obtain a modest score (0.093), higher than incorrect responses but far lower than correct ones. 
This gradient indicates that $\mathcal{L}$I distinguishes not only correctness but also the appropriateness of expressed uncertainty: it assigns high values to justified abstentions while assigning low values to hallucinations and unwarranted hesitation.  
These overall results show that $\mathcal{L}$I serves as a consistent and practical indicator of how models handle ambiguous or unanswerable questions, remaining robust across datasets, model families, and diverse prompt formulations.

\subsection{Do we really need to consider all layers instead of the final layer?}
\label{section:all_layers}

\begin{figure*}[t!]
    \centering
    \vspace{-0.8em}
    \includegraphics[width=0.9\linewidth]{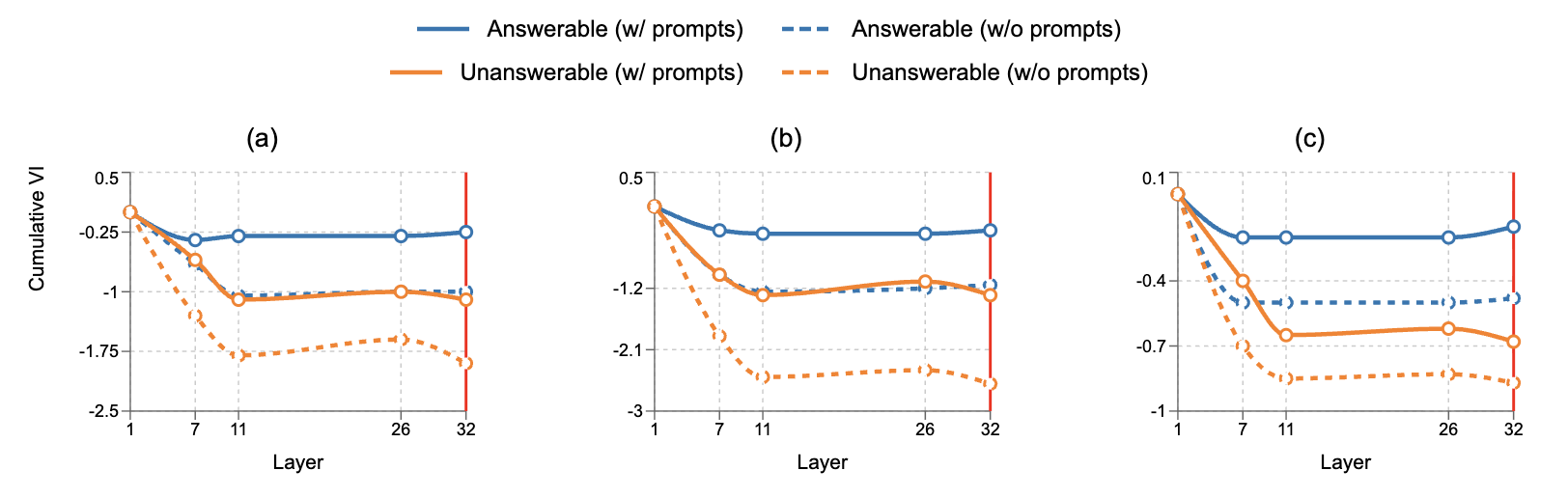}
     \vspace{-1.2em}
    \caption{Cumulative $\mathcal{V}$I, accumulating from the beginning to the layer $i$. (a) CondaQA, (b) CoQA, (c) QuAC. The $\mathcal{L}$I  scores accumulated from the beginning to the last layers (red vertical line) show the apparent, accurate differences between answerable and unanswerable questions in both settings with and without instruction prompts.}
    \label{fig:cumulative_vi}
    
\end{figure*}

\begin{table}[t!]
    \centering
    \scriptsize
    \begin{tabular}{l|cc|cc}
        \toprule
         & \multicolumn{2}{c}{\textbf{SQuAD}} & \multicolumn{2}{c}{\textbf{NQ}} \\
         &  NA & Binary & NA & Binary \\
        \midrule
        $\mathcal{L}I$ (i.e., all layers) & 74.78  & 76.42  &  55.93  &  57.26 \\ \hline
        \citet{SlobodkinGCDR23}  \\
        \hspace{2em} (1st layer) & 26.14 & 31.01 & 17.04 & 21.12 \\
        \hspace{2em} (last layer) & 51.78  & 59.42  &  55.23  &  56.26\\
        \bottomrule
    \end{tabular}
    \caption{Fact-based hallucination detection results (classification accuracy, \%). 
    NA: no instruction prompt. Binary: binary instruction prompt “Is this answerable?”. 
    }
    \label{tab:fact_hallucination}
    \vspace{-1.5em}
\end{table}

A critical question is whether it is sufficient to probe a single layer or a subset of layers, or whether information must be accumulated across the entire model depth. 
Our experiments suggest that this choice is non-trivial. 
For example, we observed that intermediate layers such as layer~6 already exhibit strong separation between answerable and unanswerable questions, raising the possibility that probing one such layer—or even aggregating only the first $k$ layers—might capture the relevant information without requiring the full $\mathcal{L}$I computation. 
However, Figure~\ref{fig:cumulative_vi} shows that signals at individual layers are not stable: while some intermediate layers appear informative, others lose or distort information before it reaches the final layer. 
As a result, relying on any single layer or partial accumulation risks missing critical dynamics. 
By contrast, $\mathcal{L}$I accumulated across all layers consistently provides a clearer and more reliable distinction, demonstrating that usable information must be tracked throughout the full network depth.

To further test whether aggregating across all layers is necessary, we evaluate fact-based hallucination detection using the probing framework of \citet{SlobodkinGCDR23}. 
As shown in Table~\ref{tab:fact_hallucination}, $\mathcal{L}$I, which accumulates information across the full model depth, consistently outperforms single-layer methods. 
On SQuAD~\citep{rajpurkar2018know}, it achieves 74.78\% and 76.42\% accuracy under no-prompt and binary-prompt settings, respectively, compared to 51.78\% and 59.42\% for last-layer probing and only 26.14\% and 31.01\% for first-layer probing. 
On NQ~\citep{kwiatkowski2019natural}, the advantage of all-layer accumulation remains, though the margins are narrower (55.93\% and 57.26\% vs. 55.23\% and 56.26\% for the last layer). 
Binary prompts improve performance across all methods, confirming the value of explicit guidance, but the gains are most pronounced for the all-layers $\mathcal{L}$I approach.  
This reinforces that aggregating usable information across all layers provides a more robust signal for hallucination detection than relying on any single layer.

\subsection{Are $\mathcal{L}$I scores computationally inexpensive?}

\begin{table*}[t!]
\centering
\scriptsize
\begin{tabular}{lccccccc}
\toprule
\textbf{Dataset} & \textbf{Context} & \textbf{Question} & $\boldsymbol{\mathcal{L}I_{\text{context + question}}}$ & $\boldsymbol{\mathcal{L}I_{\text{question}}}$ & $\boldsymbol{\mathcal{L}I_{\text{total}}}\dagger$ & \textbf{\textsc{P(True)}} & \textbf{Semantic Entropy} \\
\midrule
CoQA    & 271 words  & 5.5 words  & $1.00$ & $0.02$ & $1.02\times$ & $11\times$ & $100\times$ \\
QuAC    & 401 tokens & 6.5 tokens & $1.00$ & $0.01$ & $1.01\times$ & $11\times$ & $100\times$ \\
CondaQA & 131 tokens & 24.4 tokens & $1.00$ & $0.16$ & $1.16\times$ & $11\times$ & $100\times$ \\
\bottomrule
\end{tabular}
\vspace{-0.5em}
\caption{Computational overhead of $\mathcal{L}$I compared to \textsc{P(True)} and Semantic Entropy (SE). 
Unlike \textsc{P(True)} ($11\times$ overhead) or SE ($100\times$ overhead), $\mathcal{L}$I requires a lightweight question-only pass, $1.01$--$1.18\times$ overhead in practice.}
\vspace{-1em}
\label{tab:overhead}
\end{table*}

$\mathcal{L}$I scores are computationally inexpensive compared to other baseline methods. 
The approach requires two forward passes, one with context and one without.
However, because the second pass involves only the short question sequence, the marginal cost is negligible (Table~\ref{tab:overhead}).
According to dataset statistics, CoQA averages 271 context words and 5.5 question words \cite{ReddyCM19}, QuAC averages 401 context tokens and 6.5 question tokens \cite{ChoiHIYYCLZ18}, and CondaQA averages 131 context tokens and 24.4 question tokens \cite{Ravichander0M22}. 
As a result, the actual overhead is close to a single forward pass: $1.02\times$ on CoQA, $1.01\times$ on QuAC, and $1.16\times$ on CondaQA.  

In contrast, competing methods are far more computationally demanding. 
\textsc{P(True)} incurs about $11\times$ cost because each test query is paired with $k$ demonstrations plus the target input, yielding $(k+1)\times$ forward passes (with $k=10$, reduced from $20$ in \citet{kadavath2022language}). 
Semantic Entropy (SE) is even more expensive. It estimates uncertainty by generating 50 samples for each input, each conditioned on a 20-shot prompt \cite{farquhar2024detecting}, which results in roughly $100\times$ overhead.  

\subsection{Can calibration metrics such as ECE apply to $\mathcal{L}$I?}
\label{section:ece}

While AUROC is our primary evaluation metric, one may ask whether calibration metrics such as Expected Calibration Error (ECE) are also applicable and provide meaningful insights in this setting.
Since $\mathcal{L}$I produces scalar confidence values, they fit a logistic regression model to map them into $[0,1]$, following standard practice for ECE.
Calibration is performed on a small held-out subset of the training data, separate from the evaluation set.
We examine this on QuAC dataset with binary classification in two settings: question (un)answerability (Table~\ref{tab:ece-answerability}) and instruction-prompt ambiguity (Table~\ref{tab:ece-ambiguity}).  
For verbalized baselines, No Prompt uses per-token log-likelihoods; and Binary Prompt computes probability mass on "True" token, following \emph{Just Ask for Calibration}~\citep{tian-etal-2023-just}.

The results show consistent trends with AUROC.
First, $\mathcal{L}$I achieves lower ECE than verbalized baselines across all conditions, indicating that its scores are inherently better calibrated.  
One exception is under the binary instruction prompt with 10 calibration examples (LI 0.343 vs. Verbalized 0.312) to capture uanswerability, but it goes back to the consistent when it is trained with 100 examples.
As expected, LI shows stronger with binary than without prompt.  
In ambiguity detection, $\mathcal{L}$I maintains an advantage, reaching an ECE as low as 0.039.  
Although AUROC remains the primary evaluation metric given the advantage of $\mathcal{L}$I as parameter-free answerability signal, the ECE results highlight its reliable calibration across tasks and prompt types.  

\begin{table}[t!]
\centering
\scriptsize
\begin{tabular}{lcccc}
\toprule
\textbf{Prompt} & \textbf{\#Trainset} & \textbf{Method} & \textbf{ECE ↓} \\
\midrule
No instruction Prompt& 10  & $\mathcal{L}$I & 0.365 \\
               &     & Verbalized     & 0.451 \\
               & 100 & $\mathcal{L}$I & 0.187 \\
               &     & Verbalized     & 0.398 \\
\midrule
Binary (yes/no) & 10  & $\mathcal{L}$I & 0.343 \\
                &     & Verbalized     & 0.312 \\
                & 100 & $\mathcal{L}$I & 0.177 \\
                &     & Verbalized     & 0.276 \\
\bottomrule
\end{tabular}
\vspace{-1em}
\caption{ECE for question (un)answerability.}
\label{tab:ece-answerability}
\vspace{-0.5em}
\end{table}

\begin{table}[t!]
\centering
\scriptsize
\begin{tabular}{lcc}
\toprule
\textbf{\#Trainset} & \textbf{Method} & \textbf{ECE ↓} \\
\midrule
10   & $\mathcal{L}$I & 0.052 \\
     & Verbalized     & 0.062 \\
100  & $\mathcal{L}$I & 0.039 \\
     & Verbalized     & 0.054 \\
\bottomrule
\end{tabular}
\vspace{-0.5em}
\caption{ECE for instruction-prompt ambiguity (binary vs.\ no instruction prompt).}
\label{tab:ece-ambiguity}
\vspace{-2em}
\end{table}

\section{Conclusion}
We propose layer-wise usable information ($\mathcal{L}$I) to detect ambiguous or unanswerable questions. 
Because prior methods exclusively rely on final layers or output spaces to estimate model confidence, we argue that tracking usable information all across the layers is critical to comprehensively understand model behaviors.

\section*{Limitations}
One limitation of our method may come from its unsupervised nature. When comparing our approach to supervised methods, it may be less optimal. While supervised techniques benefit from labeled data, enabling them to learn from specific examples, our approach targets to understand large pre-trained language models in in-context question-answering tasks. Depending on their use cases and specific purposes, some may prefer supervised methods despite the associated computational costs.

\section*{Acknowledgments}
We thank Kawin Ethayarajh for his helpful feedback on $\mathcal{V}$-usable information theory. 
We thank the Oxford Applied and Theoretical Machine Learning (OATML) Group and Torr Vision Group (TVG) at the University of Oxford for their interesting discussions that have inspired this work.
YG proposed the topic and provided the early-stage feedback.
PT reviewed the work and provided later-stage feedback. 
HK led the work by proposing the methodology, designing the experiments, and writing the paper. TL reviewed the draft and helped clarify the theoretical aspects. AB contributed to the initial discussions of the work.

\bibliography{custom}

\appendix
\section{Background in Usable Information}
\label{appendix_sec:usable_info}

In this section, we explain the information-theoretic foundations for measuring model-usable information, established by~\citet{XuZSSE20} and~\citet{EthayarajhCS22}.
Consider a model family $\mathcal{V}$ that maps text input $X$ to output $Y$. The $\mathcal{V}$-usable information quantifies the amount of information a model family can extract to predict $Y$ given $X$. This metric inversely correlates with prediction difficulty. 
The lower the $\mathcal{V}$-usable information, the harder the dataset is for $\mathcal{V}$.
Consider text that is encrypted or translated into a language with more complex grammatical structures. Such transformations decrease $\mathcal{V}$-usable information, making the prediction of $Y$ given $X$ more challenging within the same model family $\mathcal{V}$.

This concept challenges traditional information theory principles, notably Shannon's mutual information~\citep{shannon} and the data processing inequality (DPI)~\citep{pippenger1988reliable}. Shannon's theory fails to account for scenarios where $X$ contains less usable information than the mutual information $I(X;Y)$ due to encryption. Similarly, DPI cannot explain how model family $\mathcal{V}$ acquires additional information through computational constraints or advanced representation learning. Two key factors demonstrate this limitation;
(1) computational constraints prevent input data from fully representing ideal world knowledge~\citep{XuZSSE20}; (2)
advanced language models can extract meaningful features from incomplete representations, achieving progressive information gains during computation~\citep{lecun2015deep, goldfeld2021sliced}.

Recent work has introduced two frameworks adopting $\mathcal{V}$-usable information to capture these phenomena.
The first framework to capture the $\mathcal{V}$-usable information is called \textbf{predictive $\mathcal{V}$-information}~\citep{XuZSSE20}. The predictive $\mathcal{V}$-information measures how much information can be extracted from X about Y when constrained to model family $\mathcal{V}$, written as $I_{\mathcal{V}}(X \rightarrow Y)$. The greater the $I_{\mathcal{V}}(X \rightarrow Y)$, the easier the dataset is for $\mathcal{V}$. While predictive $\mathcal{V}$-information provides an aggregate measure of informativeness of computational functions or dataset difficulty, \textbf{pointwise $\mathcal{V}$-information}~\citep{EthayarajhCS22} measures usable information in individual instances with respect to a given dataset distribution, written as $\text{PVI}(x\rightarrow y)$. The higher the PVI, the easier the instance is for $\mathcal{V}$, under the given
distribution.

We first define predictive conditional $\mathcal{V}$-entropy to introduce the predictive $\mathcal{V}$-information.
We follow the formal notations, defined in~\citet{XuZSSE20}:

\paragraph{Definition 2.1}  \citep{XuZSSE20} \textit{Let predictive family $\mathcal{V} \subseteq \Omega = \{ f:\mathcal{X} \cup \varnothing  \rightarrow P(\mathcal{Y}) \}$, where $X$ and $Y$ are random variables with sample space $\mathcal{X}$ and $\mathcal{Y}$, and $P(\mathcal{Y})$ be the set of all probability measures on $\mathcal{Y}$ over the Borel algebra on $\mathcal{X}$. The \textbf{predictive conditional $\mathcal{V}$-entropy} is defined as}
\begin{equation}
    H_{\mathcal{V}}(Y|X) = \text{inf}_{f \in \mathcal{V}}\mathbb{E}_{x,y \sim X,Y}[-\log_2f[X](Y)].
\end{equation}

The conditional $\mathcal{V}$-entropy is given a random variable $X$ as side information, so the function $f[X](Y)$ produces probability distributions over the output $Y$ based on the side information $X$. Suppose $\varnothing$ denote a null input that provides no information about $Y$. Note that $\varnothing \notin X$. The predictive family $\mathcal{V}$ is a subset of all possible mappings from $X$ to $P(Y)$ that satisfies \textit{optional ignorance}; whenever the $P$ predicts the outcome of $Y$, it has the option to ignore side information, $X$. That is,
\begin{equation*}
    H_{\mathcal{V}}(Y|\varnothing) = \text{inf}_{f \in \mathcal{V}}\mathbb{E}_{y \sim Y}[-\log_2f[\varnothing](Y)].
\end{equation*}
This is identical to the classic $\mathcal{V}$-entropy, denoted as $H_{\mathcal{V}}(Y)$.
We additionally specify the notation $\varnothing$ because the conditional $\mathcal{V}$-information given null input is crucial to measure how the existence of $X$ affect $\mathcal{V}$ to obtain the relevant information. 
The entropy estimation specifies the infinite functions $f$ are in $\mathcal{V}$ as~\citet{XuZSSE20} illustrate that the predictive family in theory does not take into account the computational constraints.

\paragraph{Definition 2.2}~\citep{XuZSSE20}  \textit{Let $X$, $Y$ denote random variables with sample space $\mathcal{X} \times \mathcal{Y}$, and $\mathcal{V}$ be a predictive model or function family. Then the \textbf{predictive $\mathcal{V}$-information} from $X$ to $Y$ is defined as}
\begin{equation}
    I_{\mathcal{V}}(X\rightarrow Y) = H_{\mathcal{V}}(Y|\varnothing) -  H_{\mathcal{V}}(Y|X).
\label{eq_appendix:predictive_vi}
\end{equation}

\paragraph{Definition 2.3}~\citep{EthayarajhCS22} \textit{Given random variables $X, Y$ and a predictive family $\mathcal{V}$, the \textbf{pointwise $\mathcal{V}$-information (PVI)} of an instance $(x, y)$ is}
\begin{equation}
    I_{\mathcal{V}} (x\rightarrow y) = -\log_2{f'[\varnothing](y)} +  \log_2{f'[x](y)}.
\end{equation}
\citet{EthayarajhCS22} have extended the Equation~\ref{eq_appendix:predictive_vi} to estimate the difficulty of point-wise instances for the predictive family $\mathcal{V}$. Most neural networks fine-tuned to fit label information $Y$ meet the definition of the predictive family $\mathcal{V}$ here. If $\mathcal{V}$ were, for instance, the BERT function family, then $f'[X]$ would correspond to BERT after fine-tuning on inputs that include the side information $X$, whereas $f'[\varnothing]$ would represent the same predictive family when $X$ is ignored (e.g., before fine-tuning or using only label priors).

Higher PVI means that the instance is easy for $\mathcal{V}$ while lower PVI means difficult among the given distribution. This comes from the intuition that predicting minority instances expects $\mathcal{V}$ to require more side information of $X$ to understand the instances.

\end{document}